# Decision-Level Error Attribution and Depth-Aware Safety Evaluation of Monocular RGB Human Tracking for Autonomous Robot Navigation

Abdulahi Taiwo
Department of Computer Science and Engineering
University of South Carolina
ataiwo@email.sc.edu

Chijioke Cyriacus Ekechi
Ottawa University, Ottawa, KS USA
chijiokeekechi@gmail.com

***Abstract*—Monocular RGB human tracking is attractive for robots that work near people, but its errors reach the robot through a binary stop/proceed decision that accuracy metrics do not describe. Our earlier single-participant study (prof of concept) reported that a YOLOv8-Pose, geometric-depth and Kalman-filter pipeline with a lateral-only distance rule missed no dangers yet produced frequent unnecessary stops, without being able to explain why. Here we prove that a lateral-only distance is a lower bound on planar distance, so the rule cannot miss a danger on accurate input and its false stops follow from geometry. We then introduce a two-factor Shapley attribution that splits decision errors into rule and perception shares, with the perception share further divided into missed detections and localisation error, and apply it to public pedestrian video from the ETH and UCY datasets with simulated stationary robots, a calibration/evaluation split with disjoint pedestrians, and time-block bootstrap intervals. On annotated positions from 898 pedestrians in three scenes, the lateral rule had zero missed danger and a stop precision of 5.6–31.9%, closely matching a closed-form occupancy model. With the full RGB pipeline in two scenes, a calibrated ground-plane mapping localised pedestrians with 0.59 m and 0.51 m planar RMSE, against 1.72 m and 1.39 m for the reconstructed geometric baseline and 0.81 m and 0.68 m for Depth Anything V2 with a calibrated depth map. Detection recall of 0.43 and 0.26 made missed detections the largest source of missed danger. With the ground-plane localiser, a two-sigma uncertainty margin on planar clearance missed no more danger than the lateral rule in all 8 scene–threshold cells and produced fewer unnecessary stops in 7 of them; with the geometric baseline the same margin multiplied unnecessary stops by 2.4–3.3. All robot interactions are simulated; the results are offline decision classifications, not evidence of collision avoidance.**

## I. Introduction

Mobile robots that share corridors, wards and warehouse aisles with people must decide, continuously, whether to proceed or stop [1, 2, 3]. In industrial collaborative robotics this decision is formalised as speed and separation monitoring, where the protective separation between a person and the robot drives the robot's speed [4, 5]. Low-cost monocular RGB cameras are attractive sensors for estimating that separation, but they do not measure depth, and their errors reach the robot only through a binary decision.

Perception modules are nevertheless usually reported with geometric accuracy metrics (RMSE, MAE, ADE/FDE). Our earlier study [6] evaluated a YOLOv8-Pose, geometric-depth and Kalman-filter pipeline against OptiTrack ground truth for one participant and reported a decision-level asymmetry under a 500 mm threshold: 0.0% missed-danger frames together with 65.0% unnecessary-stop frames. That study could not say where the asymmetry came from. Its own discussion named three candidates—monocular depth error, a lateral-only distance rule, and the threshold decision itself—and it used a single participant, one session, and calibration parameters estimated on the evaluated frames.

This paper asks a narrower and testable question: *when a monocular RGB human-localisation pipeline drives a stop/proceed rule, how much of the resulting decision error is caused by the decision rule's geometry, how much by localisation error, and how much by missed detections?* We answer it on public, real pedestrian video with metric ground-plane annotations, using simulated robot placements, a calibration/evaluation split that never shares a pedestrian, and time-block bootstrap intervals.

The contributions are:

1. A formal result (Proposition 1) showing that a lateral-only distance is a lower bound on planar distance, so that rule can never miss a danger that the planar rule detects and its 0% miss rate on accurate input is a property of geometry, not of perception.
2. An analysis that separates decision errors caused by the safety rule from those caused by perception, by running both rules on ground-truth and on camera-estimated positions, and that splits perception-caused missed dangers into missed detections and localisation errors.

3. An empirical study on three real scenes (3 for rule geometry, 2 with the full RGB pipeline) comparing the reconstructed geometric-depth baseline, a calibration-based ground-plane method, and Depth Anything V2 [7], evaluated by their effect on decisions, events and unnecessary stops rather than by RMSE alone.
4. Results from the earlier laboratory study are quoted only as reported there; they were not re-computed here, because the raw OptiTrack recording was not used in this work.

## II. Related Work

**Human-aware navigation.** Surveys of socially aware navigation catalogue how human motion models enter planning [1, 2], and application studies address hospital co-navigation [3], human following and guidance [8], long-term movement prediction [9] and prediction-aware obstacle avoidance [10]. These works assume a perception front end and rarely quantify how its errors change the robot's discrete decisions.

**Safety monitoring around people.** Speed and separation monitoring couples measured separation to robot motion [4, 5]. Camera-based safety systems classify operators into safety zones [11], and monocular proximity-warning systems classify workers into proximity categories and report precision, recall and F1 [12]. They evaluate a complete system; they do not separate the contribution of the zone geometry from that of the estimator.

**Monocular human localisation.** Pedestrian distance can be inferred from 2-D keypoints with learned uncertainty [13], from the ground-contact point and scene geometry [14], or from learned monocular depth such as Depth Anything V2 [7]; multi-model monocular navigation systems have been built on such cues [15, 16]. Keypoints here come from YOLOv8-Pose [17, 18] trained on COCO [19].

**Task-aware evaluation of perception.** In autonomous driving, planner-centric metrics weight detection errors by their effect on a planner [20], and planning-aware metrics have been proposed for detection and prediction [21]. Our study shares the premise that accuracy metrics are insufficient, but targets a different question: it attributes the errors of a simple, auditable stop rule to rule geometry versus perception, in a human-safety setting with monocular depth.

**Ground truth for human motion.** Marker-based motion capture is the usual reference for markerless tracking [22, 23, 24, 25]; multi-view studios [26] and robot-mounted datasets [27] broaden the setting. We use the ETH [28] and UCY [29] pedestrian recordings, distributed with homographies through OpenTraj [30], because they were the only calibrated video-plus-metric-trajectory data obtainable from our execution environment (Section V). Their ground truth is manual annotation, not motion capture.

## III. Problem Formulation

A pedestrian at ground position $\mathbf{p}(t) \in \mathbb{R}^2$ and a stationary robot modelled as a disc of radius $R$ centred at $\mathbf{r}$ have planar clearance

$$d(t) = \| \mathbf{p}(t) - \mathbf{r} \| - R. \tag{1}$$

Let $\mathbf{e}$ be the unit ground-plane vector of the camera's lateral (image-column) direction. The lateral-only rule of [6] replaces $d$ by

$$d_{\text{lat}}(t) = |(\mathbf{p}(t) - \mathbf{r})^\top \mathbf{e}| - R. \tag{2}$$

Given a threshold $\tau$, the ground-truth label of a sample is *unsafe* (positive) if $d(t) < \tau$, and a rule outputs STOP if its distance, computed from its input $\hat{\mathbf{p}}$, is below $\tau$. A false negative (FN) is an unsafe sample with PROCEED; a false positive (FP) is a safe sample with STOP. FNR = FN/(TP+FN) and FPR = FP/(FP+TN).

**Proposition 1**. *For any input* $\hat{\mathbf{p}}$, $d_{\text{lat}}(\hat{\mathbf{p}}) \le d(\hat{\mathbf{p}})$. *Hence the set of samples on which the lateral rule stops contains the set on which the planar rule stops; on the same input the lateral rule has* $\text{FN}_{\text{lat}} \le \text{FN}_{\text{pl}}$ *and* $\text{FP}_{\text{lat}} \ge \text{FP}_{\text{pl}}$. *With exact input (*$\hat{\mathbf{p}} = \mathbf{p}$*),* $\text{FN}_{\text{lat}} = 0$ *and the false positives are exactly the samples inside the strip* $|\Delta^\top \mathbf{e}| < \tau + R$ *but outside the disc* $\| \Delta \| < \tau + R$*, with* $\Delta = \mathbf{p} - \mathbf{r}$.

*Proof.* $|\Delta^\top \mathbf{e}| \le \| \Delta \| \| \mathbf{e} \| = \| \Delta \|$ (Cauchy–Schwarz), so $d_{\text{lat}} \le d$, and $d < \tau \Rightarrow d_{\text{lat}} < \tau$. The rest follows by comparing the two stop sets sample by sample. □

If pedestrian positions were spread uniformly along the camera depth direction over a length $D \gg \tau + R$, the fraction of lateral-rule stops that are justified would be the disc-to-strip area ratio $\pi(\tau + R)/(2D)$. The rule's precision therefore falls as the scene gets deeper and rises with the threshold, independently of the sensor. Real occupancy is not uniform; we report measured values.

**Error attribution.** Let $E(I, \rho)$ be an error rate (FNR or FPR) obtained with input $I \in$ {GT,RGB} and rule $\rho \in$ {lat,pl} on the same samples. The total change from the ideal system $E(\text{GT,pl})$ to the deployed one $E(\text{RGB,lat})$ is split into the two-player Shapley shares

$$\phi_{\text{rule}} = \tfrac{1}{2}\left[\left(E_{\text{GT,lat}} - E_{\text{GT,pl}}\right) + \left(E_{\text{RGB,lat}} - E_{\text{RGB,pl}}\right)\right], \tag{3}$$

$$\phi_{\text{perc}} = \tfrac{1}{2}\left[\left(E_{\text{RGB,pl}} - E_{\text{GT,pl}}\right) + \left(E_{\text{RGB,lat}} - E_{\text{GT,lat}}\right)\right], \tag{4}$$

which sum exactly to $E_{\text{RGB,lat}} - E_{\text{GT,pl}}$. Because $E_{\text{GT,pl}} = 0$ by construction, the two shares partition the deployed error. We compute them twice: on all samples (so that detection misses count as perception error) and on the samples where the pipeline produced an estimate (isolating localisation error). Under the planar rule we also split FNR into samples with no estimate and samples with a wrong estimate.

## IV. Framework and Pipeline

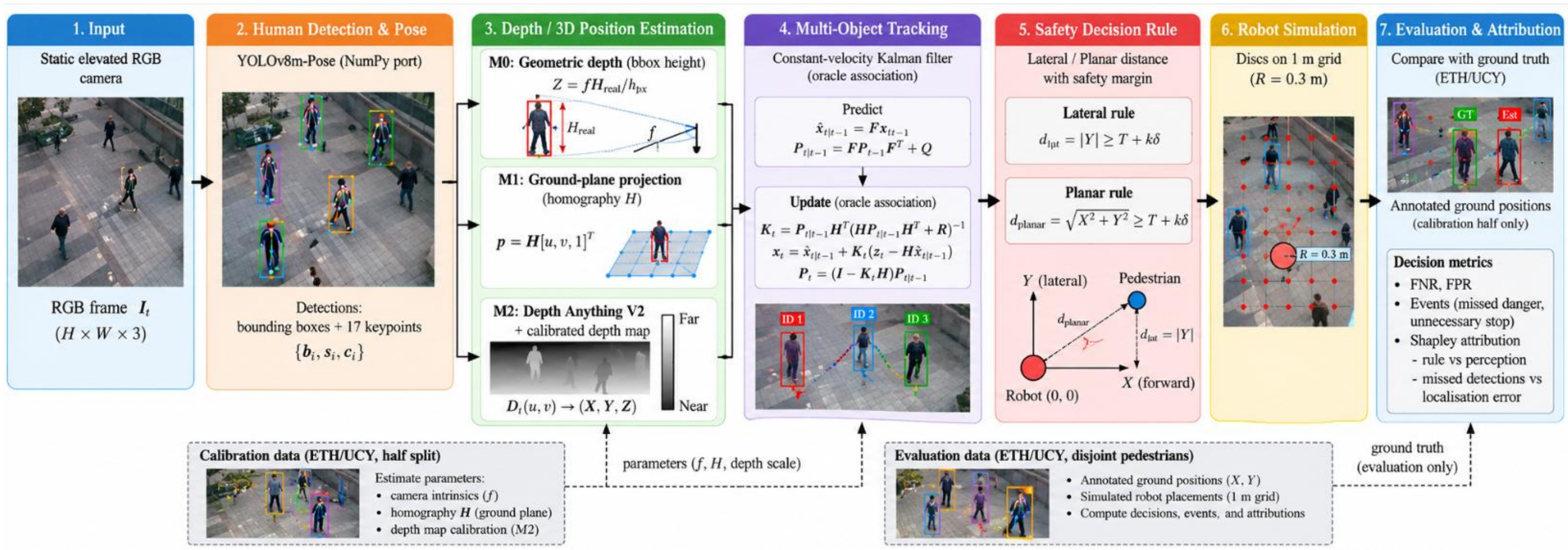

Fig. 1. Evaluated chain (conceptual diagram). Ground truth enters the estimation path only through parameters fitted on the calibration half (dashed). Robot placements are simulated. The oracle input arm of the attribution replaces the estimated positions by annotated ones.

Fig. 1 shows the evaluated chain. Every stage is fixed before evaluation-half ground truth is inspected.

**Detection and keypoints.** YOLOv8m-Pose [17] (official v8.3.0 checkpoint) runs on every annotated frame at an inference size of 1280 px with confidence threshold 0.25 and NMS IoU 0.7. Probing smaller models and the default 640 px input showed near-zero recall on these elevated views (Section VI-A). Because PyTorch could not be installed in the execution environment, we run the checkpoint through a NumPy re-implementation of the network (batch-norm folded into convolutions, DFL box decoding, the Ultralytics keypoint decoding and NMS), checked on the Ultralytics sample images.

**Localisation methods.** We compare three ways to place a detected person on the ground plane.

- *M0, geometric (reconstruction of [6]).* Depth along the optical axis from the vertical nose-to-knee pixel span $h_{\text{px}}$, $Z = fH_{\text{real}}/h_{\text{px}}$, with $H_{\text{real}} = 0.9$ m and $f$ from an assumed 65° horizontal field of view, as in the original. The lateral coordinate comes from the body-centre pixel column (mean of shoulders and hips). The original's min–max normalisation and its median-ratio depth scale both used the evaluated ground truth; we replace them by affine maps (column→lateral, $Z$ →depth) fitted on the calibration half only. Keypoints with confidence below 0.5 make the estimate unavailable.
- *M1, ground plane.* A reference pixel is mapped to the ground through the scene homography. No parameter is fitted; the reference point (ankle midpoint, nose, or box bottom) is chosen on the calibration half by median error.
- *M2, learned depth.* As M0, but the depth cue is Depth Anything V2-Small relative inverse depth [7] sampled at the ankle midpoint, mapped to the depth coordinate by $d = a/(\delta - t) + c$ fitted robustly on the calibration half. The metric variants of Depth Anything V2 could not be downloaded in our environment; the relative model needs this calibration.

**Tracking and prediction.** Each pedestrian's measurements feed a constant-velocity Kalman filter [31] on the ground plane ($\Delta t = 0.4$ s). The measurement covariance is the empirical covariance of calibration-half residuals of each method, and the process-noise intensity is chosen on the calibration half from $\{0.05, \ldots, 4\}\,\text{m}^2\text{s}^{-3}$. A track coasts through at most two missed detections. For prediction we compare the filter with raw constant-velocity extrapolation, a constant-position predictor, and constant-velocity extrapolation of the ground truth itself, which measures how predictable the motion is [32].

**Decision rules.** Lateral rule (2), planar rule (1), and an uncertainty margin rule that stops if $\hat{d} - k\hat{\sigma} < \tau$, with $\hat{\sigma}$ the filter's position standard deviation (or the calibration RMS error for unfiltered input) and $k \in \{0.5,1,2\}$. Thresholds $\tau \in \{0.30,0.50,0.75,1.00\}$ m on clearance.

## V. Datasets and Experimental Methodology

**Data.** We use seq_eth and seq_hotel from the ETH Walking Pedestrians dataset [28] and zara01 from UCY [29], obtained from the OpenTraj repository [30] (commit 97ec1d5). Each provides video from a static elevated camera, a homography from image to ground plane, and manually annotated ground positions in metres at 2.5 Hz. We verified the image–world convention by projecting annotations onto frames. Two further UCY scenes (zara02, students03) failed this check and were excluded. Table I summarises the scenes. Access to HumanEva-I, Human3.6M, JRDB and CMU Panoptic required registration or network hosts that were unavailable in our environment; these are the intended next validation sets.

**Simulated robots.** No robot appears in these recordings. We place a stationary robot disc ($R = 0.30$ m) at every node of a 1 m grid over the walkable area of each scene (2–98th percentile extent of annotated positions, keeping nodes within 1 m of some annotated position). Each (pedestrian, robot, frame) triple is one decision sample. We also report robot-level outcomes, where a robot stops if any tracked pedestrian triggers it. All decision results are offline classifications on recorded human motion with simulated robot placements; they do not demonstrate collision avoidance by a deployed robot.

**Splits and leakage control.** Within each scene, annotated frames up to the median frame form the calibration half; later frames form the evaluation half, restricted to pedestrian IDs that never appear in the calibration half. All fitted parameters (affine maps, reference-point choice, Kalman noise) use the calibration half only. Detections are associated with annotated pedestrians in image space by the Hungarian algorithm (distance between the projected annotation and the box centre, normalised by box height, gate 0.8). Tracking therefore uses oracle data association, which isolates localisation from identity errors.

**Events and uncertainty.** An unsafe event is a maximal run of consecutive unsafe samples for one (pedestrian, robot) pair; it is missed if no STOP occurs within it. An unnecessary-stop episode is a maximal run of consecutive false positives. Video frames are strongly correlated, so we never treat frames as independent: intervals for decision metrics come from a bootstrap over contiguous 10 s time blocks [33] (1000 resamples), and intervals for localisation and prediction errors from a bootstrap over pedestrians. With few blocks per scene these intervals are approximate, and we make no population-level significance claims.

TABLE I
SCENES USED. ANNOTATIONS ARE MANUAL GROUND-PLANE POSITIONS AT 2.5 HZ. ROBOT PLACEMENTS ARE SIMULATED.

| Scene | Duration (s) | People | Annot. | Robots | Eval. people | Det. recall |
|---|---|---|---|---|---|---|
| ETH | 773 | 360 | 8908 | 130 | – | not run† |
| Hotel | 722 | 390 | 6544 | 78 | 187 | 0.26 |
| Zara01 | 360 | 148 | 5024 | 99 | 78 | 0.43 |

† YOLOv8m-Pose recall was 0.09 on a 6-frame probe of this near-nadir view; ETH is used only with ground-truth input.

## VI. EXPERIMENTAL RESULTS

### A. Detection on elevated views

On the evaluation halves, YOLOv8m-Pose at 1280 px recovered 0.43 of annotated pedestrians in Zara01 and 0.26 in Hotel (Table I; calibration halves 0.47 and 0.28). A preliminary probe on four frames per scene found at most five detections per model and scene at the default 640 px input, and on a six-frame probe at 1280 px YOLOv8m-Pose reached a recall of 0.09 in the near-nadir ETH view, so ETH enters only the ground-truth experiment (probe outputs in the replication package). Fig. 2 shows typical frames. Detection misses, not localisation, are therefore the first-order perception error in these views, and Section VII quantifies their effect on decisions.

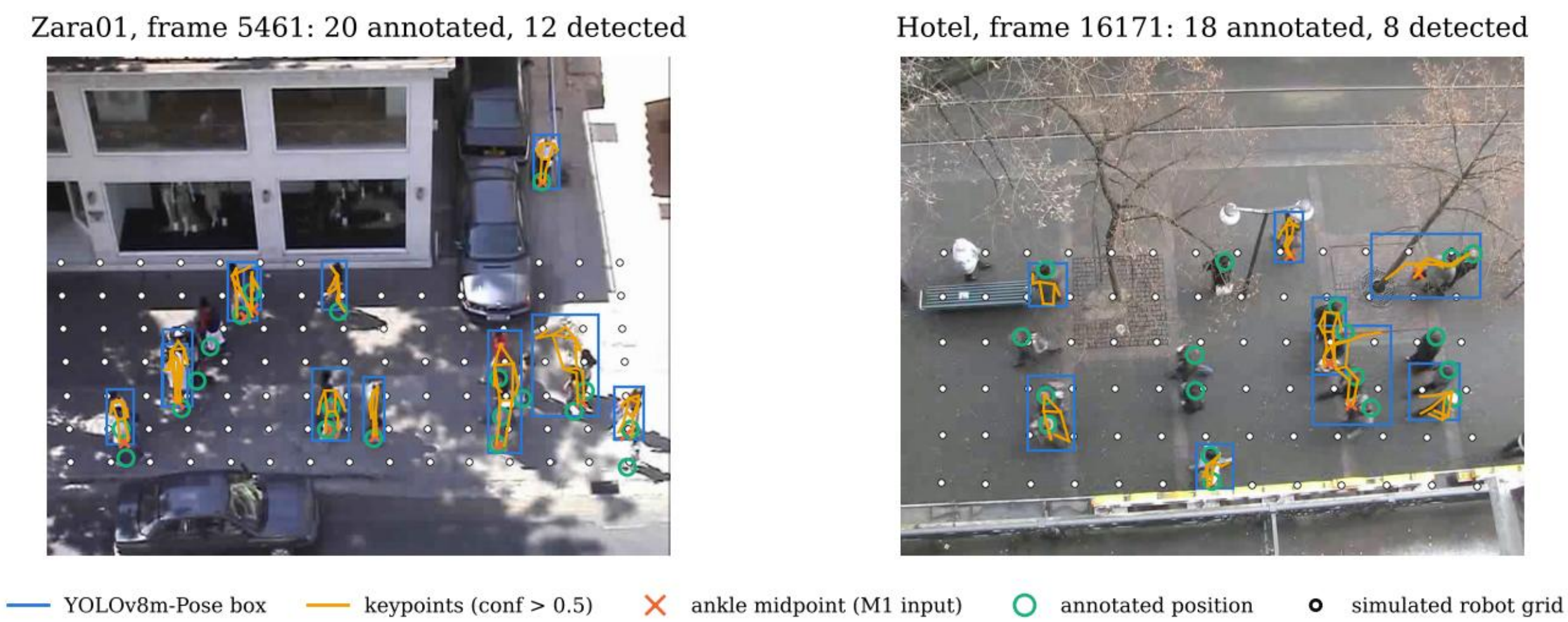

Fig. 2. Evaluation-half frames from the two RGB scenes with YOLOv8m-Pose detections, keypoints, ankle midpoints (input of M1), annotated positions, and the simulated robot grid. Source: ETH and UCY video via OpenTraj; detections computed in this study.

### B. Rule geometry on annotated positions

Fed with annotated positions, the lateral rule produced no missed danger in any of the 12 scene–threshold cells (73026 unsafe events in total), exactly as Proposition 1 requires (Table II). Its stops were nevertheless mostly unnecessary: precision ranged from 5.6% to 31.9%. At $\tau = 0.5$ m precision was 7.5% in ETH, 20.0% in Hotel and 18.8% in Zara01, against 7.5%, 21.4% and 20.9% predicted by the uniform-occupancy model of Section III. At the robot level the lateral rule kept a robot stopped while no pedestrian was within the threshold for 22–52% of the time (Fig. 3). Perception plays no part in these numbers.

TABLE II

LATERAL RULE FED WITH ANNOTATED POSITIONS (ALL FRAMES). THE PLANAR RULE IS EXACT BY CONSTRUCTION (FNR = FPR = 0). BRACKETS: 95% TIME-BLOCK BOOTSTRAP INTERVALS.

| Scene | $\tau$ (m) | Events | FNR | FPR (%) | Prec. (%) | Model (%) | Robot unn. (%) |
|---|---|---|---|---|---|---|---|
| ETH | 0.30 | 5483 | 0.0 | 14.0 [13.7, 14.3] | 5.6 [5.5, 5.7] | 5.6 | 44.1 |
| | 0.50 | 7661 | 0.0 | 18.4 [17.9, 18.8] | 7.5 [7.3, 7.6] | 7.5 | 48.2 |
| | 0.75 | 10247 | 0.0 | 23.5 [23.0, 24.1] | 9.8 [9.6, 10.0] | 9.9 | 51.4 |
| | 1.00 | 12842 | 0.0 | 28.6 [27.9, 29.3] | 12.1 [11.8, 12.3] | 12.2 | 52.4 |
| Hotel | 0.30 | 3147 | 0.0 | 7.7 [7.5, 7.9] | 14.8 [14.1, 15.3] | 16.1 | 27.4 |
| | 0.50 | 4390 | 0.0 | 9.4 [9.2, 9.6] | 20.0 [19.3, 20.6] | 21.4 | 29.7 |
| | 0.75 | 5977 | 0.0 | 11.9 [11.6, 12.3] | 25.5 [24.6, 26.3] | 28.1 | 31.4 |
| | 1.00 | 7523 | 0.0 | 13.5 [13.1, 13.8] | 31.9 [31.0, 32.6] | 34.8 | 30.5 |
| Zara01 | 0.30 | 2553 | 0.0 | 6.8 [6.7, 6.9] | 14.0 [13.7, 14.3] | 15.7 | 22.3 |
| | 0.50 | 3401 | 0.0 | 8.5 [8.4, 8.6] | 18.8 [18.4, 19.2] | 20.9 | 24.8 |
| | 0.75 | 4430 | 0.0 | 10.6 [10.5, 10.7] | 24.3 [23.9, 24.7] | 27.5 | 26.6 |
| | 1.00 | 5372 | 0.0 | 12.3 [12.1, 12.5] | 29.6 [29.0, 30.1] | 34.0 | 27.1 |

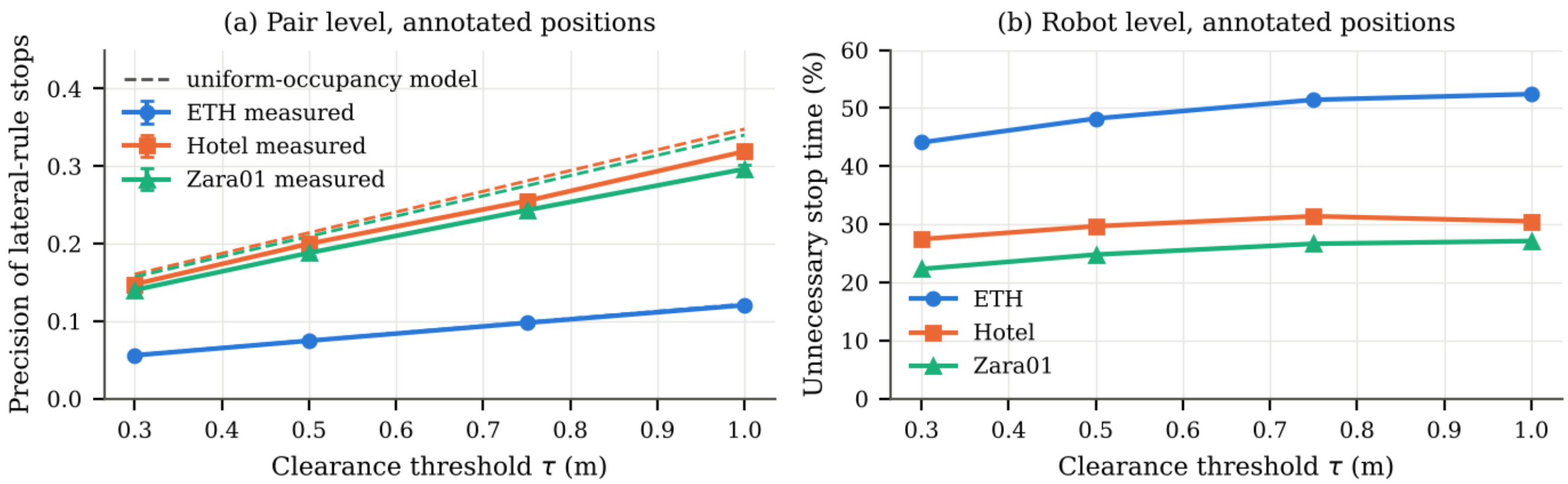

Fig. 3. Lateral rule on annotated positions (no perception). (a) Precision of stops, with 95% time-block bootstrap intervals (narrower than the markers) and the uniform-occupancy prediction $\pi(\tau + R)/(2D)$ (dashed; for ETH it coincides with the measured line). FNR is zero in every case. (b) Share of robot-time spent stopped while no annotated pedestrian was within $\tau$.

### *C. Localisation*

Table III and Fig. 4 compare the localisation methods on pedestrians unseen during calibration. The calibrated ground-plane method M1 had the lowest planar error in both scenes (Zara01 0.59 m, Hotel 0.51 m), against 1.72 m and 1.39 m for the reconstructed geometric baseline M0 and 0.81 m and 0.68 m for Depth Anything V2 with a calibrated depth map (M2). Lateral errors were similar across methods (0.20–0.22 m in Zara01, 0.24–0.26 m in Hotel); the methods differ in depth. M0 also produced estimates for fewer samples (26% of annotated evaluation samples in Zara01 versus 40% for M1), because it needs a confident nose and both knees. The M1 reference point chosen on the calibration half was the ankle midpoint in Zara01 and the nose in Hotel: Hotel annotations mark heads, and its dataset positions are those head pixels mapped through the ground homography, so mapping the nose the same way is the annotation-consistent choice. It inherits the parallax of that convention, which is part of the ground-truth limitation discussed in Section IX. Requiring a confident nose also lowered M1's coverage in Hotel (16% of annotated evaluation samples).

TABLE III

LOCALISATION ERROR ON THE EVALUATION HALF (MATCHED DETECTIONS; PEDESTRIANS DISJOINT FROM CALIBRATION). BRACKETS: 95% BOOTSTRAP OVER PEDESTRIANS. COVERAGE: SHARE OF ANNOTATED EVALUATION SAMPLES WITH AN ESTIMATE.

| Scene | Method | Coverage (%) | $n$ | Lateral RMSE (m) | Depth RMSE (m) | Depth bias (m) | Planar RMSE (m) |
|---|---|---|---|---|---|---|---|
| Zara01 | M0 geometric | 25.8 | 667 | 0.20 [0.17, 0.23] | 1.71 [1.44, 2.03] | 0.69 | 1.72 [1.46, 2.04] |
| | M1 ground plane | 40.1 | 1039 | 0.22 [0.20, 0.25] | 0.55 [0.48, 0.62] | -0.29 | 0.59 [0.52, 0.66] |
| | M2 DA-V2 + calib. | 40.1 | 1039 | 0.20 [0.18, 0.23] | 0.78 [0.69, 0.87] | 0.28 | 0.81 [0.72, 0.90] |
| Hotel | M0 geometric | 11.7 | 398 | 0.25 [0.20, 0.29] | 1.37 [1.19, 1.53] | 0.12 | 1.39 [1.21, 1.55] |
| | M1 ground plane | 15.8 | 537 | 0.26 [0.23, 0.30] | 0.44 [0.38, 0.49] | 0.20 | 0.51 [0.46, 0.57] |
| | M2 DA-V2 + calib. | 16.5 | 564 | 0.24 [0.19, 0.28] | 0.64 [0.59, 0.68] | 0.12 | 0.68 [0.63, 0.72] |

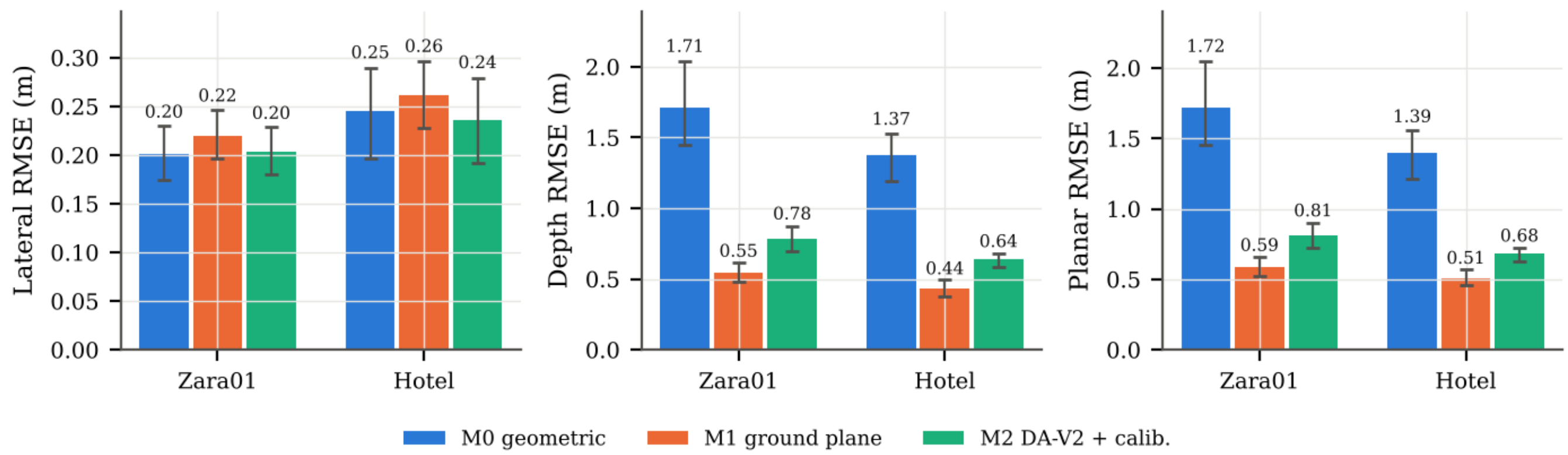

Fig. 4. Localisation RMSE on the evaluation half with 95% bootstrap intervals over pedestrians.

### *D. Decisions*

Table IV lists decision outcomes at $\tau = 0.5$ m with Kalman-filtered input. With M1 input and the planar rule, the missed-danger rate was 56.7% (Zara01) and 77.5% (Hotel), with precision 64.5% and 63.8%. The lateral rule on the same input lowered FNR to 38.8% and 70.6% but cut precision to 17.4% and 16.8%. At the robot level, with M1 input in Zara01, the lateral rule left the robot stopped unnecessarily for 20.6% of the time with a robot-level FNR of 13.0%; the planar rule gave 1.1% and 40.6%, and the planar rule with a $2\hat{\sigma}$ margin 11.7% and 15.1% (Hotel: 14.6/51.2, 1.1/71.4 and 14.0/46.7%).

TABLE IV

DECISION OUTCOMES AT $\tau = 0.5$ M ON THE EVALUATION HALF, KALMAN-FILTERED INPUT, ALL ANNOTATED SAMPLES (NO ESTIMATE ⇒ PROCEED). BRACKETS: 95% TIME-BLOCK BOOTSTRAP. UNN. STOP: SHARE OF PAIR-TIME STOPPED WHILE SAFE.

| Scene | Input | Rule | TP | FN | FP | FNR (%) | FPR (%) | Prec. (%) | Missed events | Unn. stop (%) |
|---|---|---|---|---|---|---|---|---|---|---|
| Zara01 | GT | lateral | 4981 | 0 | 21267 | 0.0 [0.0, 0.0] | 8.46 | 19.0 | 0/1778 | 8.29 |
| | GT | planar | 4981 | 0 | 0 | 0.0 [0.0, 0.0] | 0.00 | 100.0 | 0/1778 | 0.00 |
| | M0 geometric | lateral | 2274 | 2707 | 10728 | 54.3 [49.7, 58.3] | 4.27 | 17.5 | 767/1778 | 4.18 |
| | M0 geometric | planar | 483 | 4498 | 1566 | 90.3 [87.7, 92.4] | 0.62 | 23.6 | 1550/1778 | 0.61 |
| | M0 geometric | planar $+1\hat{\sigma}$ | 1720 | 3261 | 12521 | 65.5 [60.1, 70.0] | 4.98 | 12.1 | 956/1778 | 4.88 |
| | M0 geometric | planar $+2\hat{\sigma}$ | 2417 | 2564 | 31449 | 51.5 [45.9, 56.0] | 12.51 | 7.1 | 695/1778 | 12.27 |
| | M1 ground plane | lateral | 3046 | 1935 | 14448 | 38.8 [35.1, 41.7] | 5.75 | 17.4 | 503/1778 | 5.63 |
| | M1 ground plane | planar | 2157 | 2824 | 1187 | 56.7 [51.5, 61.1] | 0.47 | 64.5 | 806/1778 | 0.46 |
| | M1 ground plane | planar $+1\hat{\sigma}$ | 2826 | 2155 | 3947 | 43.3 [38.5, 47.1] | 1.57 | 41.7 | 589/1778 | 1.54 |
| | M1 ground plane | planar $+2\hat{\sigma}$ | 3092 | 1889 | 8226 | 37.9 [33.6, 41.3] | 3.27 | 27.3 | 508/1778 | 3.21 |
| | M2 DA-V2 + calib. | lateral | 3051 | 1930 | 14290 | 38.7 [35.1, 41.6] | 5.68 | 17.6 | 500/1778 | 5.57 |
| | M2 DA-V2 + calib. | planar | 1720 | 3261 | 1583 | 65.5 [62.6, 68.6] | 0.63 | 52.1 | 984/1778 | 0.62 |
| | M2 DA-V2 + calib. | planar $+1\hat{\sigma}$ | 2665 | 2316 | 5299 | 46.5 [42.8, 50.2] | 2.11 | 33.5 | 614/1778 | 2.07 |
| | M2 DA-V2 + calib. | planar $+2\hat{\sigma}$ | 3093 | 1888 | 11326 | 37.9 [34.2, 41.0] | 4.50 | 21.5 | 495/1778 | 4.42 |
| Hotel | GT | lateral | 6136 | 0 | 24266 | 0.0 [0.0, 0.0] | 9.34 | 20.2 | 0/2470 | 9.13 |
| | GT | planar | 6136 | 0 | 0 | 0.0 [0.0, 0.0] | 0.00 | 100.0 | 0/2470 | 0.00 |
| | M0 geometric | lateral | 1354 | 4782 | 7118 | 77.9 [74.7, 80.6] | 2.74 | 16.0 | 1766/2470 | 2.68 |
| | M0 geometric | planar | 580 | 5556 | 1518 | 90.5 [88.4, 92.2] | 0.58 | 27.6 | 2110/2470 | 0.57 |
| | M0 geometric | planar $+1\hat{\sigma}$ | 1327 | 4809 | 8261 | 78.4 [74.7, 81.2] | 3.18 | 13.8 | 1696/2470 | 3.11 |
| | M0 geometric | planar $+2\hat{\sigma}$ | 1656 | 4480 | 20399 | 73.0 [69.5, 76.0] | 7.85 | 7.5 | 1568/2470 | 7.67 |
| | M1 ground plane | lateral | 1805 | 4331 | 8971 | 70.6 [67.3, 73.1] | 3.45 | 16.8 | 1530/2470 | 3.37 |
| | M1 ground plane | planar | 1380 | 4756 | 783 | 77.5 [74.3, 80.0] | 0.30 | 63.8 | 1764/2470 | 0.29 |
| | M1 ground plane | planar $+1\hat{\sigma}$ | 2002 | 4134 | 3755 | 67.4 [63.9, 70.2] | 1.45 | 34.8 | 1407/2470 | 1.41 |
| | M1 ground plane | planar $+2\hat{\sigma}$ | 2137 | 3999 | 8765 | 65.2 [61.8, 68.1] | 3.37 | 19.6 | 1337/2470 | 3.30 |
| | M2 DA-V2 + calib. | lateral | 1807 | 4329 | 8903 | 70.6 [66.5, 74.0] | 3.43 | 16.9 | 1534/2470 | 3.35 |
| | M2 DA-V2 + calib. | planar | 1108 | 5028 | 1005 | 81.9 [79.7, 84.0] | 0.39 | 52.4 | 1877/2470 | 0.38 |
| | M2 DA-V2 + calib. | planar $+1\hat{\sigma}$ | 1935 | 4201 | 4896 | 68.5 [64.8, 71.7] | 1.88 | 28.3 | 1469/2470 | 1.84 |
| | M2 DA-V2 + calib. | planar $+2\hat{\sigma}$ | 2124 | 4012 | 11558 | 65.4 [61.1, 69.2] | 4.45 | 15.5 | 1385/2470 | 4.35 |

### *E. Prediction*

With M1 input, the Kalman filter's planar prediction RMSE grew from 0.64 m at 0.4 s to 1.46 m at 2.0 s in Zara01 and from 0.62 to 2.06 m in Hotel (Table V, Fig. 5). It was the best of the three measurement-based predictors at every horizon. Raw constant-velocity extrapolation was worse than holding the last position at every horizon in Zara01 and only slightly better than it beyond 0.4 s in Hotel. Constant-velocity extrapolation of the annotations themselves stayed at 0.05–0.39 m (Zara01) and 0.12–0.50 m (Hotel). Prediction error here is dominated by measurement error rather than by the motion model, which argues against a learned predictor: it would be trained on a few hundred short tracks and could not remove measurement noise that the filter already absorbs. We therefore did not train one.

TABLE V
PLANAR PREDICTION ERROR (M, RMSE) FROM M1 INPUT ON THE EVALUATION HALF. "CV ON GT" EXTRAPOLATES THE ANNOTATIONS THEMSELVES.

| Scene | Predictor | 0.4s | 0.8s | 1.2s | 2.0s |
|---|---|---|---|---|---|
| Zara01 | Kalman CV | 0.64 | 0.80 | 1.00 | 1.46 |
| | raw CV | 0.88 | 1.29 | 1.74 | 2.63 |
| | const. position | 0.74 | 1.08 | 1.49 | 2.36 |
| | CV on GT | 0.05 | 0.12 | 0.20 | 0.39 |
| | $n$ (KF) | 1011 | 990 | 962 | 904 |
| Hotel | Kalman CV | 0.62 | 0.92 | 1.28 | 2.06 |
| | raw CV | 0.70 | 0.97 | 1.33 | 2.12 |
| | const. position | 0.68 | 1.12 | 1.62 | 2.65 |
| | CV on GT | 0.12 | 0.21 | 0.29 | 0.50 |
| | $n$ (KF) | 482 | 459 | 429 | 369 |

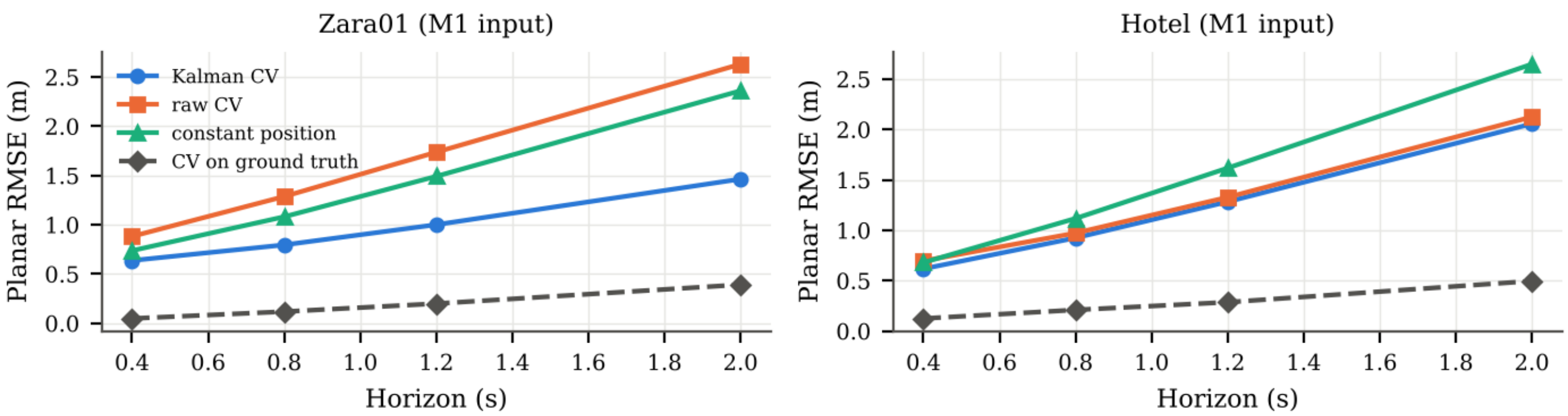

Fig. 5. Planar prediction RMSE versus horizon (M1 input, evaluation half). "CV on ground truth" bounds what a constant-velocity model could achieve with perfect measurements.

### *F. Computational cost*

On a 2-core CPU without GPU, the NumPy YOLOv8m-Pose port took 4.3 s per 720×576 frame at 1280 px and Depth Anything V2-Small (ONNX Runtime, 518×518) 0.9 s; localisation, filtering and decisions are negligible. These timings reflect the constrained environment, not the models' achievable speed on robot hardware.

## VII. Decision-Level Error Analysis

Table VI and Fig. 6 decompose the deployed error (RGB input, Kalman filter, lateral rule) into rule and perception shares. Three patterns hold for every method in both scenes.

**The rule trades misses for stops.** The rule share of FNR is negative and the rule share of FPR positive: replacing the planar rule by the lateral one hides misses by stopping more often. With M1 in Zara01 the rule share was -8.9 pp of FNR and +6.87 pp of FPR.

**Perception drives missed danger.** The perception share of FNR was 47.8 pp for M1 in Zara01 (74.0 pp in Hotel). Under the planar rule, 33.0 pp of the 56.7% FNR in Zara01 came from samples with no estimate and 23.7 pp from wrong estimates (Hotel: 64.2 and 13.3 pp of 77.5%). Restricted to samples with an estimate, M1's FNR fell to 35.4% (planar) and 8.7% (lateral) in Zara01.

**Perception slightly lowers FPR, for the wrong reason.** The perception share of FPR is negative because missed pedestrians cannot trigger stops. It is not evidence of a well-calibrated estimator.

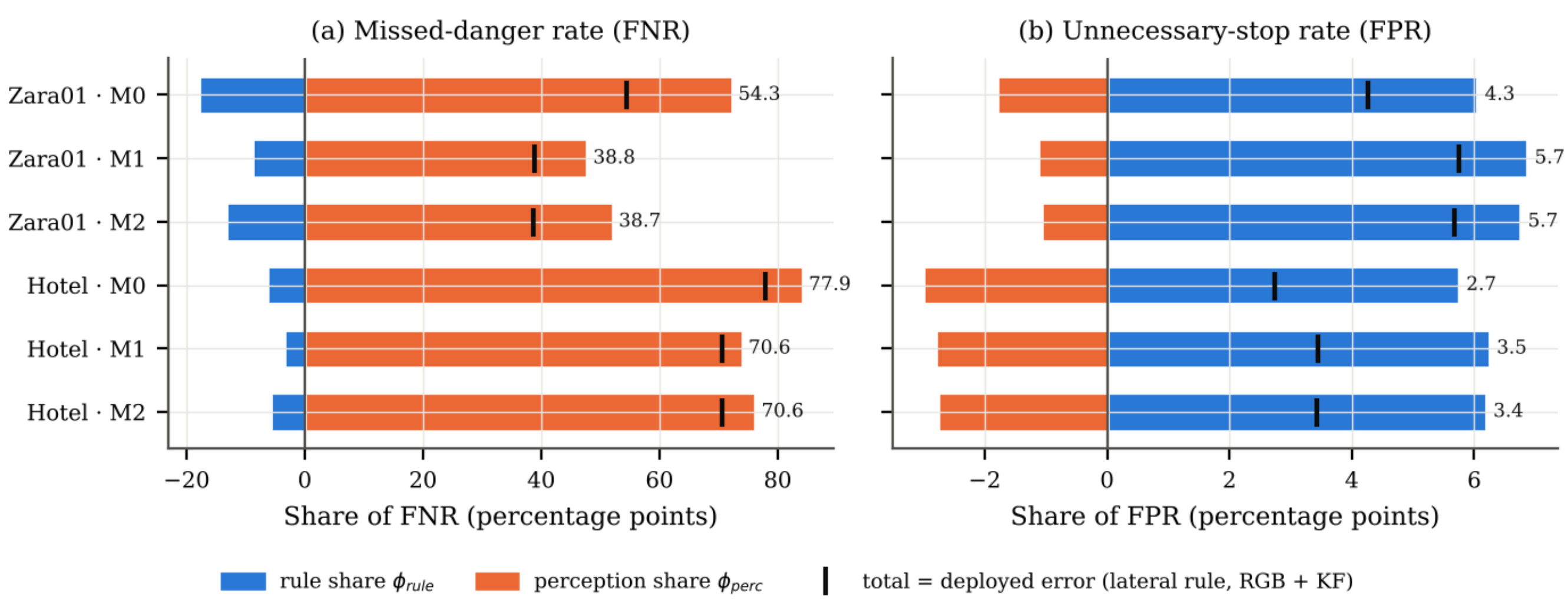

Fig. 6. Shapley attribution of the deployed error at $\tau = 0.5$ m (all annotated evaluation samples). Bars: rule and perception shares (they sum to the total, the deployed system's error rate).

TABLE VI
TWO-FACTOR SHAPLEY ATTRIBUTION AT $\tau = 0.5$ M (KALMAN-FILTERED INPUT), IN PERCENTAGE POINTS. "ALL": EVERY ANNOTATED SAMPLE (MISSES COUNT AS PERCEPTION ERROR). "DETECTED": SAMPLES WITH AN ESTIMATE ONLY. BRACKETS: 95% TIME-BLOCK BOOTSTRAP.

| | | | FNR (pp) | | | FPR (pp) | | | FNR planar split (pp) | |
|---|---|---|---|---|---|---|---|---|---|---|
| Scene | Method | Samples | Total | $\phi_{\text{rule}}$ | $\phi_{\text{perc}}$ | Total | $\phi_{\text{rule}}$ | $\phi_{\text{perc}}$ | No est. | Loc. |
| Zara01 | M0 geometric | all | 54.3 | -18.0 [-19.4, -16.8] | 72.3 [68.9, 75.3] | 4.27 | 6.05 | -1.78 | 49.1 | 41.2 |
| | | detected | 10.3 | -35.3 [-37.3, -33.3] | 45.6 [44.1, 47.2] | 8.56 | 7.86 | 0.70 | | |
| | M1 ground plane | all | 38.8 | -8.9 [-10.0, -7.7] | 47.8 [43.5, 51.6] | 5.75 | 6.87 | -1.12 | 33.0 | 23.7 |
| | | detected | 8.7 | -13.3 [-15.4, -11.2] | 22.0 [19.5, 24.3] | 8.70 | 8.23 | 0.46 | | |
| | M2 DA-V2 + calib. | all | 38.7 | -13.4 [-14.9, -12.3] | 52.1 [49.2, 54.9] | 5.68 | 6.76 | -1.07 | 33.0 | 32.5 |
| | | detected | 8.6 | -19.9 [-21.8, -18.6] | 28.5 [26.7, 30.8] | 8.60 | 8.07 | 0.53 | | |
| Hotel | M0 geometric | all | 77.9 | -6.3 [-7.2, -5.6] | 84.2 [81.4, 86.2] | 2.74 | 5.75 | -3.01 | 72.4 | 18.1 |
| | | detected | 20.0 | -22.9 [-25.1, -20.8] | 42.9 [40.1, 45.3] | 10.33 | 8.85 | 1.48 | | |
| | M1 ground plane | all | 70.6 | -3.5 [-3.9, -3.0] | 74.0 [70.7, 76.5] | 3.45 | 6.25 | -2.79 | 64.2 | 13.3 |
| | | detected | 17.9 | -9.7 [-10.9, -8.4] | 27.5 [24.3, 30.8] | 10.09 | 9.36 | 0.73 | | |
| | M2 DA-V2 + calib. | all | 70.6 | -5.7 [-7.1, -4.6] | 76.2 [73.1, 78.8] | 3.43 | 6.19 | -2.76 | 64.7 | 17.3 |
| | | detected | 16.7 | -16.1 [-18.5, -14.1] | 32.8 [30.2, 35.3] | 9.86 | 9.06 | 0.80 | | |

**Operating points.** Fig. 7 plots FNR against FPR over the four thresholds. With the ground-plane localiser M1 ($\hat{\sigma} \approx$ 0.30 m before filtering), the planar rule with a $2\hat{\sigma}$ margin had an FNR no higher than the lateral rule in 8 of 8 scene–threshold cells and a lower FPR in 7 of 8; the exception was Hotel at $\tau =$ 0.3 m. With M2 the comparison was mixed (6 and 4 of 8). With M0, whose calibration error is about 1.24 m (Zara01) and 0.98 m (Hotel), the margin lowered FNR in every cell but multiplied FPR by 2.4–3.3. An explicit margin is therefore an efficient way to buy conservatism only when the estimator's uncertainty is small; with a poor depth estimate, no rule recovers what the sensor did not measure. At the robot level (Table IV text, Section VI), the $2\hat{\sigma}$ margin with M1 cut unnecessary stop time by about 40% in Zara01 relative to the lateral rule at a slightly higher miss rate, and in Hotel it was marginally better on both.

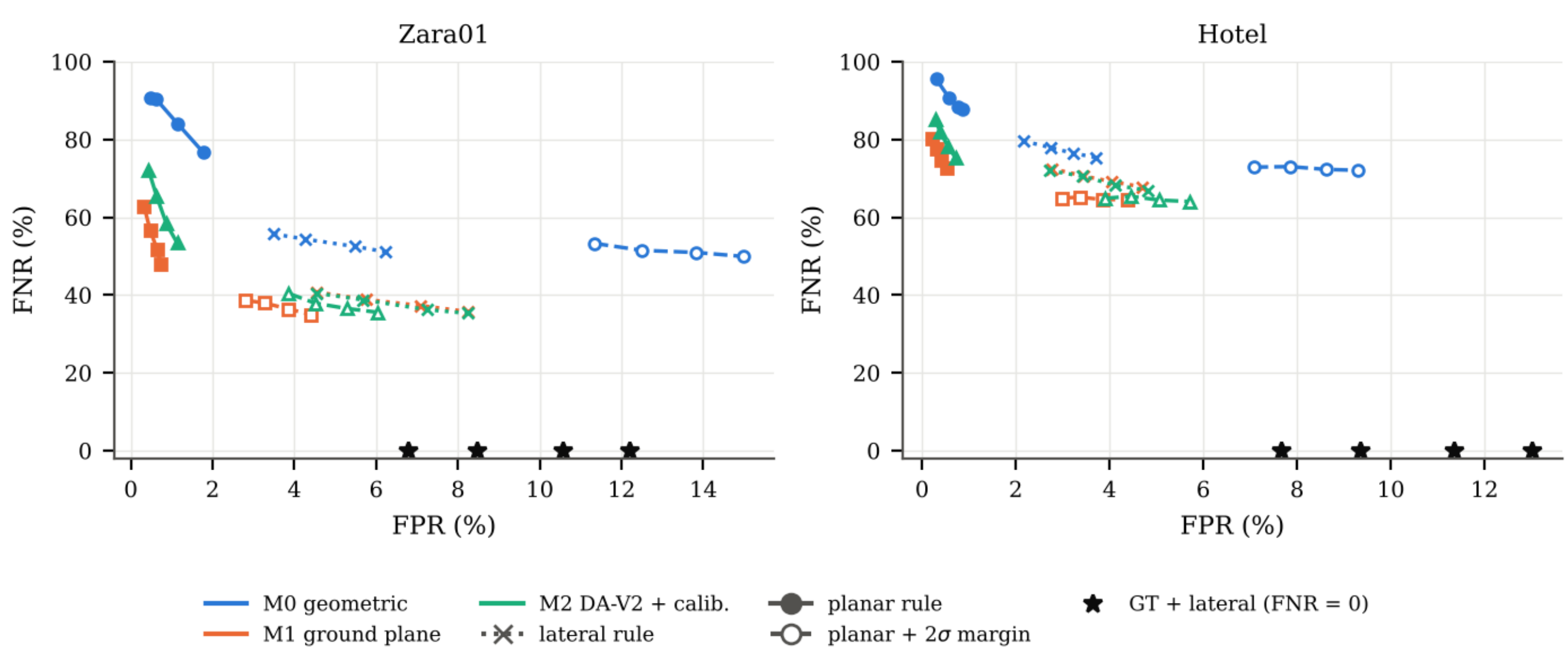

Fig. 7. Missed-danger rate versus unnecessary-stop rate across thresholds $\tau \in \{0.3, 0.5, 0.75, 1.0\}$ m (pair level, Kalman-filtered input, evaluation half). Stars: annotated positions with the lateral rule.

## VIII. DISCUSSION AND PRACTICAL IMPLICATIONS

The earlier laboratory study [6] reported 0.0% false negatives and 65.0% false-positive frames for one participant. Proposition 1 and Table II show that a zero miss rate is the expected behaviour of a lateral-only rule whenever lateral localisation is accurate, and that a large false-stop burden follows from the geometry alone. The lateral rule is a conservative bound, not a sensor property. The laboratory values themselves are not re-derived here and are not comparable numerically: that study used one participant, an OptiTrack reference, a 120 Hz overhead camera, no robot radius and different frame denominators.

A single accuracy number hides three different failure sources with opposite effects on safety. We recommend that monocular safety pipelines report (i) FNR and FPR (or precision) with their denominators and intervals over time blocks, (ii) event-level missed dangers, (iii) the share of FNR caused by missing estimates, and (iv) the same metrics with the decision rule fed by ground truth, which separates rule design from perception.

In these views a calibrated ground-plane mapping of a body keypoint beat both the reconstructed anthropometric depth and a calibrated learned relative-depth model, and it needs no fitted parameters beyond the scene homography. The largest remaining safety lever was detection coverage: a stop rule cannot react to a person the detector never sees. For a good localiser, an uncertainty margin is a better way to buy conservatism than discarding depth, because the margin is explicit, tunable, and shrinks as the estimator improves; for a poor one it mostly adds false stops.

## IX. Limitations

**Ground truth.** ETH and UCY positions are manual annotations at 2.5 Hz projected through homographies, not motion capture. Their own error (annotation point, timing, homography) is unquantified and is included in every error we report; Fig. 2 shows visible offsets between some annotations and feet. **Scenes and subjects.** Two scenes were usable with the full RGB pipeline, and one more with annotated input. They are outdoor pedestrian areas viewed from above, not robot viewpoints. Pedestrians differ between calibration and evaluation, but cameras do not; a cross-camera test was not possible for M0 and M2, whose maps are scene-specific. **Simulated robots.** Robots are stationary discs on a grid; no robot was present, moving, or reacting. Results are offline classifications of recorded motion and do not show collision avoidance. **Unannotated people.** ETH and UCY skip some pedestrians near image borders; we therefore score only annotated pedestrians and associate detections to them with ground truth, which excludes false detections of unannotated people and makes tracking oracle-associated. **Detector configuration.** We used COCO-trained YOLOv8m-Pose without fine-tuning; detectors trained for top-down views would change the detection share

## X. Conclusion

Evaluating monocular RGB human tracking through the stop/proceed decisions it drives exposes failure modes that accuracy metrics hide. A lateral-only distance rule never misses a danger when fed accurate positions, but most of its stops are unnecessary, by geometry; on real video, missed detections then dominate the missed dangers. A two-factor attribution separates these effects with the same data, and, for a ground-plane localiser, an explicit uncertainty margin on planar clearance offers a better trade-off than discarding depth. The next step is to repeat the protocol with motion-capture ground truth from multiple calibrated views and with a real robot viewpoint.